# A Realtime Autonomous Robot Navigation Framework for Human like High-level Interaction and Task Planning in Global Dynamic Environment

Sung-Hyeon Joo, Sumaira Manzoor, Yuri Goncalves Rocha, Hyun-Uk Lee and Tae-Yong Kuc
College of Information and Communication Engineering
Sungkyunkwan University
Suwon, Republic of Korea
sh.joo@skku.edu, sumaira11@skku.edu, yurirocha@skku.edu, zlshvl36@skku.edu. tykuc@skku.edu

*Abstract*— In this paper, we present a framework for real-time autonomous robot navigation based on cloud and on-demand databases to address two major issues of human-like robot interaction and task planning in global dynamic environment, which is not known a priori. Our framework contributes to make human-like brain GPS mapping system for robot using spatial information and performs 3D visual semantic SLAM for independent robot navigation. We accomplish the feat by separating robot's memory system into Long-Term Memory (LTM) and Short-Term Memory (STM). We also form robot's behavior and knowledge system by linking these memories to Autonomous Navigation Module (ANM), Learning Module (LM), and Behavior Planner Module (BPM). The proposed framework is assessed through simulation using ROS-based Gazebo-simulated mobile robot, RGB-D camera (3D sensor) and a laser range finder (2D sensor) in 3D model of realistic indoor environment. Simulation corroborates the substantial practical merit of our proposed framework.

## I. INTRODUCTION

High-level interaction, planning and ability of independent navigation are essential functions for a robot to perform human-like tasks in dynamic wide-area environment. Traditional methods of operating robots and performing tasks based on user commands or sensor information rely on pre-planned task planning and simple calibration by low-level sensor information feedback. These are not suitable for performing human-like tasks who identifies changes in the working environment or understands the inherent meaning of a situation. For making robots to perform high-level tasks, human learning and cognitive skills that enhance their environmental adaptation, task planning and performance by utilizing past work experiences and acquired knowledge should be granted. This means a new method of modeling the environment and operating the robot is required.

In this paper, our goal is to propose a real-time autonomous robot navigation framework for human-like robot interaction and task performance in dynamic environment**.** The framework, based on human outstanding spatial representation and the ability to create spatial maps, includes advanced mapping systems, autonomous navigation, task planning and learning technologies.

We have contributed using cutting-edge technologies based on the human ability with the following characteristics:
- Human-like brain GPS mapping system with excellent spatial scalability
- Acquiring and utilizing rich information such as geometric properties, physical properties and properties of spatial components to build spatial maps.
- Learning and application of spatial information through the semantic-episodic memory

The rest of the paper is organized as follows: In Section II, we provide the literature review. In Section III, we explain the key features of our proposed framework, state-of-the-art environment, behavior, knowledge and map models that imitate human abilities. This section also provides detailed description of our framework's components. Section IV presents the significant effects of our proposed framework in simulated environment. Finally, we conclude our work in Section V.

## II. RELATED WORK

In this section, we review the existing work in literature that has inspired us to propose a framework, presented in this paper. Environmental modeling technology that utilizes various sensors to extract semantical information from objects and spaces is being researched along with semantic mapping technology that tries to converge semantic information into topologic and metric maps. However, there are limitations to perform high-level tasks. As the research in this field is still in a relatively early stage, which currently uses only simple information extracted from the environment. Research on making 3D object models and a knowledge database on the cloud is underway. These are attempts to integrate the cloud DB with robot task planning and semantic-based mapping. However, obtaining real-time scalability for wide-area dynamic spaces and various pragmatic tasks is unknown.

Our proposed framework combines aspects of two research areas: Semantic SLAM and Task Planning. The remaining part of this section gives a brief literature overview of these two areas.

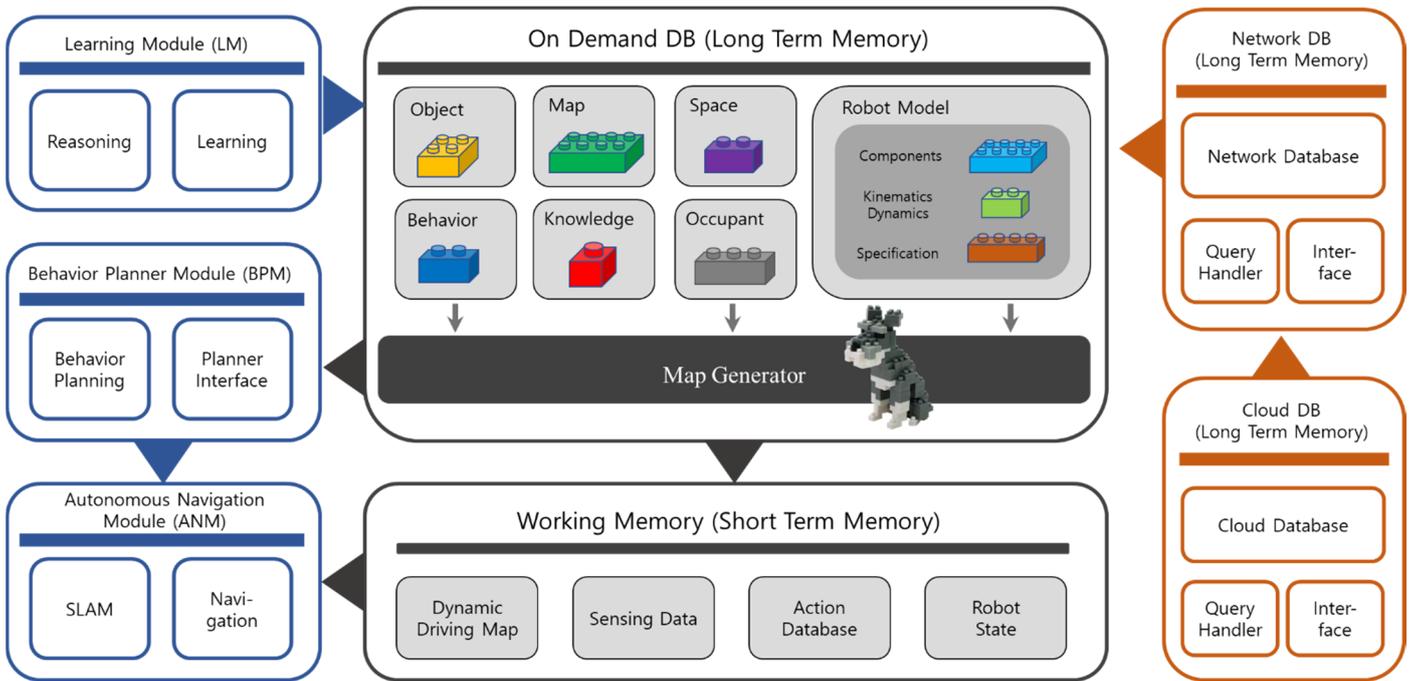

Fig. 1. Overview of the proposed framework: A robot-centric on-demand database and a working memory are used for autonomous navigation, behavior planner, and learning. The robot can download database from network or cloud if necessary.

### A. Semantic SLAM

Simultaneous Localization and Mapping (SLAM), aims to self-localize a robot and estimate a model of its environment by acquiring a map from sensory information. For effective interaction of a robot with its environment, it is necessary to obtain semantic understanding of the 3D map constructed from a visual SLAM system. Several research efforts have been made recently for investigating the approaches to conduct visual SLAM on semantic level [1], [2].

Most studies of Visual SLAM rely on sparse, semi-dense, or dense cloud points, depending on the image alignment methods. Traditional feature extraction algorithms only exploits limited feature points that contain lines segments, image corners and blobs, such as classic ORBSLAM [3] and MonoSLAM [4]. Therefore, they are not appropriate for 3D semantic mapping of environment due to limited feature points.

Current approaches make use of direct dense SLAM system [5], such as surfel-based ElasticFusion [6] and Dense Visual SLAM [7] for better extraction of available image information to avoid calculation cost of features and obtain higher pose accuracy. Direct image alignment through these dense methods is suitable for object recognition tasks using RGB-D camera.

Most recent approaches using deep CNNs provide efficient methods for sematic segmentation of an image to understand the environment with more robust knowledge. The first seminal idea of pixel-wise prediction on semantic segmentation from supervised pre-training using fully convolutional network (FCN) was introduced by [8]. Then, a completely different approach [9] using deep deconvolution networks to generate dense and accurate semantic segmentation masks and overcome the limitation of fixed-size receptive field in the FCN [8]. DeepLab system [10] achieved high performance on semantic segmentation by extracting dense features using atrous convolution and up-sampled filters. The major challenge to achieve reasonable accuracy through these deep CNN models was high computational cost. To overcome this issue, Xception [11] used more efficient model parameters and gained the robust performance without increasing capacity. Recently, MobileNets [12] introduced an efficient convolution neural network architecture to boost the performance of semantic segmentation by building low latency models for mobile vision applications.

### B. Task Planning

In an autonomous robot, task planning should be performed efficiently using a sequence of high-level actions for accomplishing a given task in real-time. Its efficiency depends on data storage capacity and reduced computational time. Different types of knowledge are required to encode into the planner to perform a task in appropriate way. These include
- Causal knowledge that is related to the effects of the robot's actions.
- World knowledge that contains information about different objects, their relations and properties.

Knowledge about the world environment and its structure is encoded in form of a map. Now, researchers are more focused on semantic maps to extend the robots capabilities for performing the fundamental tasks such as navigation, exploration of environment and manipulation [13]. The sheer volume of data is generated by vision sensors to accomplish these tasks in dynamic environment, but it requires fast processing and storage of thousands of images. An approach to reduce the storage and computational requirements of VSLAM by offloading the processing to a cloud is described in [14]. In a recent study [15], a cloud-enabled framework for robots known as "Robot-cloud" is presented to perform computationally intensive tasks of navigation, map building and path planning by preparing knowledge repository to share vision sensor's information on the cloud.

### III. FRAMEWORK

Our proposed framework minimizes the dependency of a cloud database for robot's independent and real-time performance. Block diagram in Fig. 1 illustrates our proposed framework. This framework, consisting of robot, network and cloud, is designed with a robot-centric architecture. The core of this framework is robot's memory system separated by Long-Term Memory (LTM) and Short-Term Memory (STM), based on the human memory system. The LTM is comprised of the robot mounted on-demand database that stores environmental information, behavior, knowledge, and map data. The STM which is used as a working memory stores information obtained from sensors and the dynamic driving map for self-driving. These memories are organically linked to Autonomous Navigation Module (ANM), Learning Module (LM), and `Behavior Planner Module (BPM) to form robot's behavior and knowledge system. Network and cloud databases are LTM, stored in the network and cloud respectively. These databases complement the limited storage capacity of robots through an interface with the on-demand database.

#### A. On-demand Database

Environmental and behavioral information, knowledge and map data are stored in the on-demand database. The robot retrieves the data available in the on-demand database to perform its mission. After obtaining the cloud data, the robot uses it to plan the behavioral actions and builds a dynamic driving map according to the given mission. The robot also takes advantage of the newly obtained information to modify the on-demand database in parallel for the execution of given task. Whenever additional information about the environment or task is needed, it can be merged with the robot's current knowledge by downloading it from the network and cloud databases.

A robot-mounted on-demand environment database is designed to model rich properties of space elements, in conjunction with a multi-modal sensing information for representing a variety of environmental information. The on-demand environment database includes objects, spaces, robot models, occupants and inherent knowledge about each of its elements to express high-level environmental information in a human-like fashion.

An ontology-based advanced spatial representation is used to define each component of environment database as shown in Fig. 2. It consists of a set of explicit models, implicit models and symbols to express various geometric characteristics, as well as material and relationship information. Moreover, the explicit model defines all the geometrical and physical information that can be retrieved by the sensors, while the implicit model describes the intrinsic relation between objects, spaces and occupants. The symbolic model defines any element in a language-oriented way.

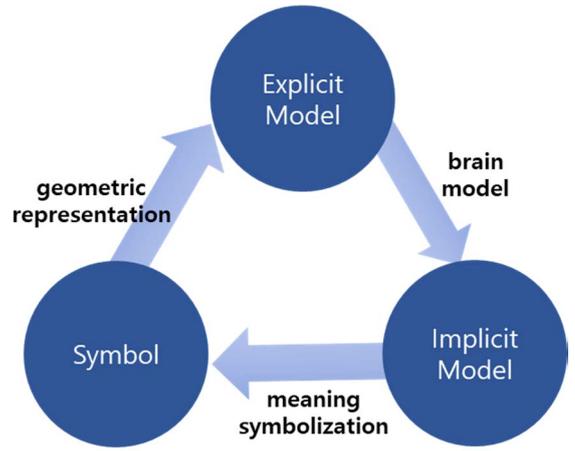

Fig. 2. Element modeling concept for Environment database

#### B. Semantic Map Generation

We generate different levels of semantic maps according to the specifications of robot′s sensors and the given task. Fig. 3 shows a map system based on the findings of cognitive science, with a class hierarchy composed of characteristic, structure, form and symmetric properties.

In our proposed approach, the hierarchical mapping system, its effectiveness and real-time capabilities of human brain map model are combined as a framework that provides a semantic map of dynamic environment with broad coverage capability and enables the robot to quickly adapt to wide area dynamic spaces.

#### C. Working Flow

When the robot is assigned a task, each module operates sequentially. Map Generator (MG) generates an optimal semantic-episodic map for the robot's resources, conditions, and mission environment based on the on-demand database. BPM plans behavior by matching the generated map with behavior database. According to the planned behavior sequence, ANM builds an optimal dynamic driving map, considering both the operating space and mission to perform real-time autonomous navigation. New information obtained during driving is learned by reasoning in the LM.

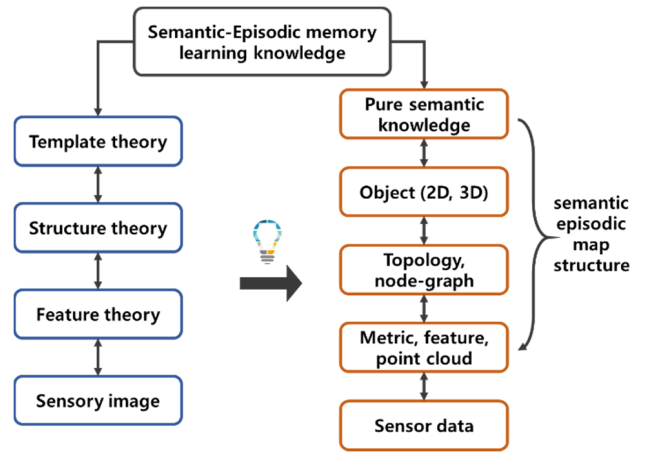

Fig. 3. Cognitive science-based map system hierarchy

## IV. SIMULATION

We demonstrate the usability of our proposed framework by designing a simulation environment of a convention center which includes static objects, moving actors and a four-wheeled robot, using Gazebo simulator on ROS platform, as shown in Fig. 4. To test robot's components and its behavior in different surroundings, SDF (Simulated Description Format) based description of the environment and URDF (Unified Robot Description File) based description of the robot are integrated with Gazebo. SDF is an XML format for describing objects and environments of a world in Gazebo simulator. URDF is an XML format for representing a robot model which can be efficiently integrated with both, ROS and Gazebo. Accurate models of the simulated robot and the exhibition hall working environment are designed. The simulated robot includes an RGB-D camera (3D sensor) and a laser range finder (2D sensor).

The simulation environment represents a complete convention center, with two different exhibition halls and a main lobby. Several exhibition booths are placed inside one hall, while the other hall has a stage and an agglomeration of people.

The environment information is stored in the on-demand database according to the proposed framework. Objects and environmental data are stored using the symbols, explicit and implicit models. The explicit model is further divided into 2D sensor and 3D sensor models. Simulation using RGB-D camera and a laser range finder shows the feasibility of our proposed framework for performing autonomous navigation tasks and semantic SLAM in dynamic environment.

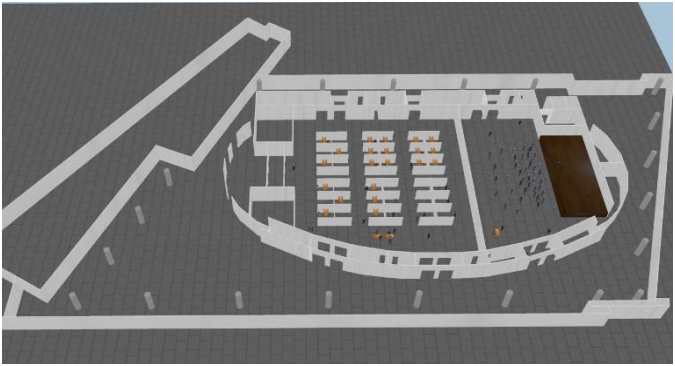

Fig. 4. Simulation environment

## V. CONCLUSION AND FUTURE WORK

We have proposed the self-driving technology framework to perform high-level interactions and human-like tasks. The framework suggests, how robots can perform tasks flexibly in global dynamic environments. In addition, robot customized on-demand environment mapping technology enables different types of robots to be used in various fields for providing personal and professional services. We present the use of the framework through simulation and plan to apply the framework to various environments in future research.

## ACKNOWLEDGMENT

This research was supported by Korea Evaluation Institute of Industrial Technology (KEIT) funded by the Ministry of Trade, Industry & Energy (MOTIE) (No. 1415158956)